\documentclass[letterpaper]{article} 
\usepackage{aaai23}  
\usepackage{times}  
\usepackage{helvet}  
\usepackage{courier}  
\usepackage[hyphens]{url}  
\usepackage{graphicx} 
\usepackage{natbib}  
\usepackage{caption} 
\usepackage{algorithm}
\usepackage{algorithmic}
\usepackage{amsmath}
\allowdisplaybreaks[4]
\usepackage{booktabs}
\usepackage{multirow}

\usepackage{newfloat}
\usepackage{listings}
\usepackage{amssymb}
\DeclareCaptionStyle{ruled}{labelfont=normalfont,labelsep=colon,strut=off} 
\floatstyle{ruled}
\newfloat{listing}{tb}{lst}{}
\floatname{listing}{Listing}
\title{Grouped Knowledge Distillation for Deep Face Recognition}
\author{
    Weisong Zhao \textsuperscript{\rm1,3\equalcontrib},
    Xiangyu Zhu \textsuperscript{\rm2,4\equalcontrib},
    Kaiwen Guo \textsuperscript{\rm2},
    Xiao-Yu Zhang \textsuperscript{\rm1,3\thanks{Corresponding Author.}},
    Zhen Lei \textsuperscript{\rm2,4,5} 
}
\affiliations{
    \textsuperscript{\rm 1}Institute of Information Engineering, Chinese Academy of Sciences, Beijing, China\\
\textsuperscript{\rm 2} CBSR\&NLPR, Institute of Automation, Chinese Academy of Sciences, Beijing, China\\
\textsuperscript{\rm 3} School of Cyber Security, University of Chinese Academy of Sciences, Beijing, China\\
\textsuperscript{\rm 4} School of Artificial Intelligence, University of Chinese Academy of Sciences, Beijing, China\\
\textsuperscript{\rm 5} Centre for Artificial Intelligence and Robotics, Hong Kong Institute of Science \& Innovation,\\
Chinese Academy of Sciences, Hong Kong, China\\

   \{zhaoweisong, zhangxiaoyu\}@iie.ac.cn, guokaiwen1991@gmail.com, \{xiangyu.zhu, zlei\}@nlpr.ia.ac.cn
}

\usepackage{bibentry}

\begin{document}

\maketitle

\begin{abstract}
Compared with the feature-based distillation methods, logits distillation can liberalize the requirements of consistent feature dimension between teacher and student networks, while the performance is deemed inferior in face recognition. One major challenge is that the light-weight student network has difficulty fitting the target logits due to its low model capacity, which is attributed to the significant number of identities in face recognition. Therefore, we seek to probe the target logits to extract the primary knowledge related to face identity, and discard the others, to make the distillation more achievable for the student network. Specifically, there is a tail group with near-zero values in the prediction, containing minor knowledge for distillation. To provide a clear perspective of its impact, we first partition the logits into two groups, i.e., Primary Group and Secondary Group, according to the cumulative probability of the softened prediction. Then, we reorganize the Knowledge Distillation (KD) loss of grouped logits into three parts, i.e., Primary-KD, Secondary-KD, and Binary-KD. Primary-KD refers to distilling the primary knowledge from the teacher, Secondary-KD aims to refine minor knowledge but increases the difficulty of distillation, and Binary-KD ensures the consistency of knowledge distribution between teacher and student. We experimentally found that (1) Primary-KD and Binary-KD are indispensable for KD, and (2) Secondary-KD is the culprit restricting KD at the bottleneck. Therefore, we propose a Grouped Knowledge Distillation (GKD) that retains the Primary-KD and Binary-KD but omits Secondary-KD in the ultimate KD loss calculation. Extensive experimental results on popular face recognition benchmarks demonstrate the superiority of proposed GKD over state-of-the-art methods.
\end{abstract}

\section{Introduction}

\begin{figure}[htb]
\centering
\includegraphics[width=\columnwidth]{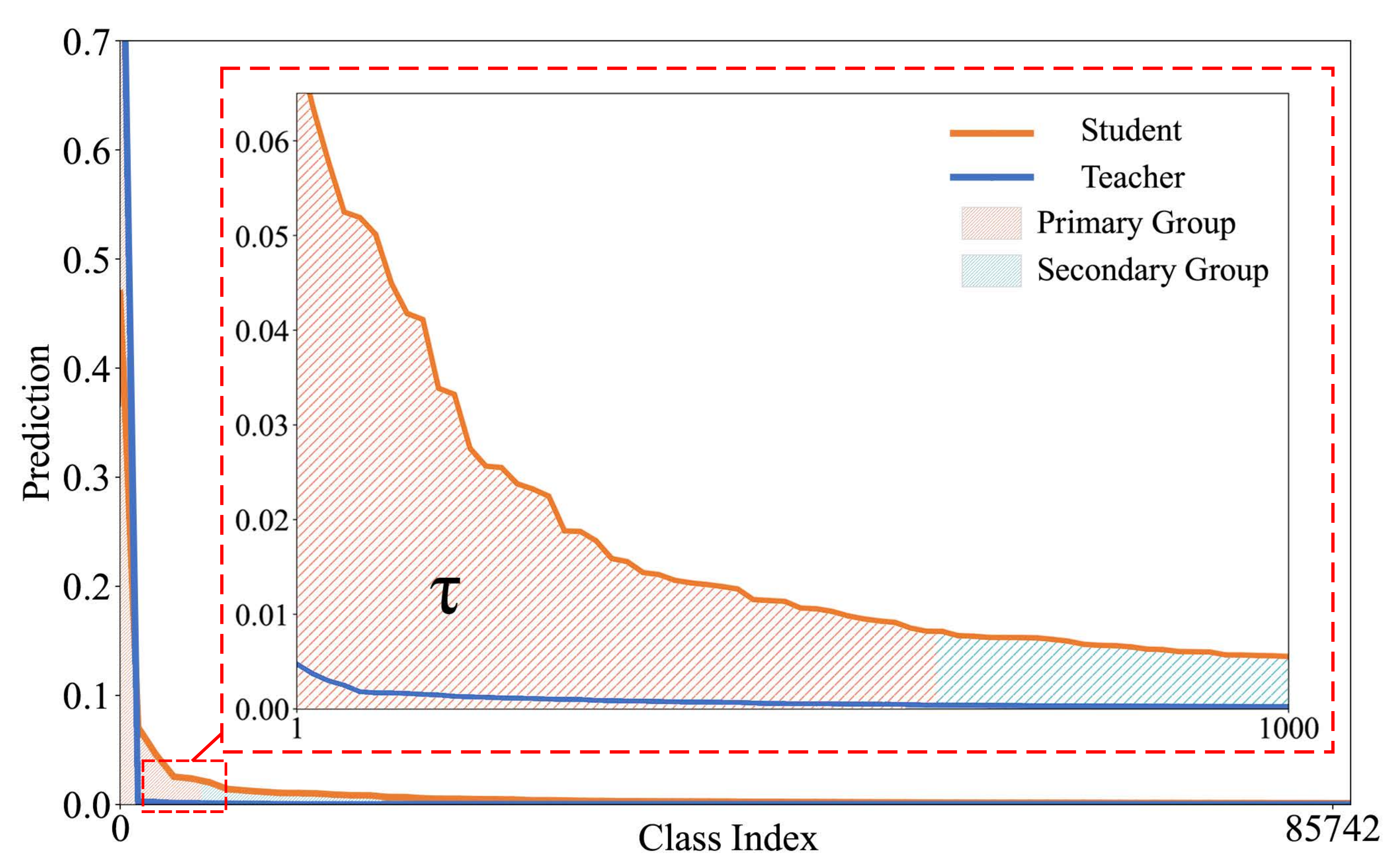}
\caption{The distribution of the predictions (softened probabilities) of student and teacher networks trained from scratch. The analysis is conducted on MS1MV2 \cite{arcface} dataset. The predictions are visualized based on a randomly selected sample. There is a long tail group that has several near-zero values, containing minor knowledge for distillation. We partition the original logits into two groups, i.e., primary group (pink) and secondary group (blue), via a cumulative probability threshold $\tau$ of ranked student prediction. Best viewed in color.}
\label{fig1}
\end{figure}
Face recognition has achieved great success in various application domains \cite{liziqing,leizhen}. However, a large number of light-weight yet discriminative face recognition models are required due to the development of mobile and edge devices. An intuitive solution is to optimize the neural network architectures for mobile devices, e.g., MobileFaceNet \cite{mobilefacenet} and MobileNetV3 \cite{mobilenetv3}. However, discriminative networks always benefit from a large model capacity, which will introduce more computational and storage costs. Given a heavy teacher network, Knowledge Distillation (KD) aims to improve the accuracy of light-weight networks, where the knowledge from a heavy teacher network is transferred to the light-weight student network.

The idea of KD \cite{kd} was first introduced to transfer knowledge by reducing the Kullback-Leibler (KL) divergence between the prediction probabilities of the teacher and the student networks. In the past decade, the research attention has been drawn to conducting instance-wise constraints on the activation of the hidden layers, e.g., FitNet \cite{fitnet} distills knowledge from deep features of intermediate layers. However, such feature-based methods require the teacher and student networks to share the same representation space, which is unrealistic for student networks with low model capacities \cite{ekd}. Additionally, extra computational and storage usage (e.g., additional network modules and identical feature dimension requirements) are introduced for distilling deep features.

Unlike feature-based distillation, logits distillation does not require the student to mimic the teacher’s representation space, but rather to preserve the high semantic consistency with the teacher, which can liberalize the requirements of consistent feature dimension between teacher and student networks. Unfortunately, the performance of logits distillation is inferior in large-scale face recognition. One major challenge is that the light-weight student network has difficulty fitting the target logits due to its low model capacity, which is attributed to the great number of identities in face recognition. Therefore, we seek to probe the target logits to extract the primary knowledge related to face identity and discard the others, to make the distillation more achievable for the student network. Specifically, we can see from Fig. \ref{fig1} that there is a tail group with near-zero values in the softened prediction, which contains minor knowledge for distillation and wastes the learning capabilities of student models.

To provide a clear perspective for the impact of the tail group, we partition the output logits into two groups via the cumulative probability threshold of the student's softened prediction, i.e., primary group and secondary group. We argue that secondary group contains minor knowledge and can not be covered by the student network with low model capacity. To more distinctly embody the role of secondary group in the classic KD, we reorganize the classical KD loss into three parts, i.e., Primary-KD, Secondary-KD, and Binary-KD. Primary-KD distills the most discriminative knowledge from the teacher, Secondary-KD aims at distilling minor knowledge embedded in the tail group, and Binary-KD ensures the consistency of knowledge distribution between teacher and student. The extensive experiments prove that (1) Primary-KD and Binary-KD are indispensable elements for KD, and (2) Secondary-KD is the culprit restricting KD at the bottleneck. Therefore, we propose a Grouped Knowledge Distillation (GKD), which remains Primary-KD and Binary-KD in the ultimate loss calculation. The extensive comparisons with current SOTA methods on several popular face benchmarks demonstrate the superiority of the proposed GKD.

Overall, the contributions of this paper are summarized as follows:
\begin{itemize}
	 \item We propose to find an achievable distillation method for the student network to bridge the performance gap between teacher and student models. Specifically, we introduce the grouped logits to partition the logits into two groups, i.e., primary group and secondary group, via cumulative probability threshold of corresponding softened prediction.
    \item Given the grouped logits, we propose reorganizing the classical KD loss into three parts, i.e., Primary-KD, Secondary-KD, and Binary-KD. Specifically, we experimentally analyze and prove the individual impacts of these three components; that is, Primary-KD and Binary-KD are indispensable for KD, and Secondary-KD is the culprit restricting KD at the bottleneck. On the basis of the reformulation of KD loss, we propose a Grouped Knowledge Distillation that retains the Primary-KD and Binary-KD in the KD loss computation.
    \item The extensive experiments on popular face recognition benchmarks demonstrate the superiority of the proposed GKD over the state-of-the-art methods.
\end{itemize}

\section{Related Work}
\paragraph{Loss Functions.}
There are two types of loss functions applied for face recognition. The first is the verification loss function. Contrastive loss \cite{contrastive1,contrastive2} optimizes pairwise Euclidean distance in feature space. Triplet loss \cite{facenet,tripletnet} makes up triplets to separate the positive pair from the negative pair by a distance margin. The second is the softmax-based loss function, which is mostly adopted by current SOTA deep face recognition methods.  To learn angularly discriminative features, the SphereFace \cite{sphereface} was proposed by introducing the angular SoftMax function (i.e., A-SoftMax), which imposes discriminative constraints on a hypersphere manifold. Also, the CosFace \cite{cosface} with a large margin cosine loss was proposed further to maximize the decision margin in the angular space. The additive
angular margin loss ArcFace loss was designed to obtain highly discriminative features for FR \cite{arcface}. CurricularFace \cite{curricularface} embeds the idea of curriculum learning into the loss function. Softmthatsed loss functions combined with heavy neural networks are demonstrated to obtain satisfactory performance \cite{arcface}, but they cannot be well deployed in the light-weight network in the mobile sciences \cite{lightweight}. Specifically, a performance gap exists between heavy and mobile models, which requires the application of knowledge distillation. 

\paragraph{Knowledge Distillation.} Knowledge distillation was first proposed by Hinton {\em et al.} \cite{kd}, which refers to a model compression method to transfer the knowledge of a heavy teacher model to a light-weight student network. Later variants of distillation strategies are proposed to utilize diverse information from the teacher model, which can be divided into two types, i.e., logits distillation and feature distillation. As for the feature distillation, FitNet \cite{fitnet} bridges the middle layers of the student and teacher networks and adopted $L_2$ loss to further supervised the output of the student. ShrinkTeaNet \cite{shrinkteanet} proposes minimizing the angle between teacher and student embedding vectors. TripletDistillation \cite{tripletdistillation} improves the triplet loss with dynamic margins by utilizing the similarity structures among different identities in the teacher network. MarginDistillation \cite{margindistillation} utilizes class prototypes from the teacher network for the student network. Additionally, SP \cite{sp} adopt the pairwise similarities of the outputs. EKD introduces the evaluation-oriented method to optimize the student model’s critical relations \cite{ekd}. Most feature distillation strategies could perform better than logits distillation but introduce more computational and storage costs. %

Different from feature-based distillation methods, logits distillation does not require the student to mimic the teacher’s representation space, but rather to preserve the high semantic consistency with the teacher. The classical KD \cite{kd} proposes to minimize the Kullback-Leibler Divergence of softened class probabilities between the teacher and student. Besides, DML \cite{dml} suggests mutual learning to train multiple networks simultaneously. Mirzadeh {\em et al.} \cite{takd} introduce multi-step knowledge distillation, which employs an intermediate-sized network to bridge the gap between the student and the teacher. Li {\em et al.} propose a nested collaborative learning structure \cite{jun}. DKD \cite{dkd} firstly proposes to decouple the classical KD Loss into target class knowledge distillation and non-target class knowledge distillation. However, we argue that DKD has inferior performance over large-scale face recognition because considerable logits will dramatically increase the difficulty of distillation.
\begin{figure*}[htb]
\centering
\includegraphics[width=0.95\textwidth]{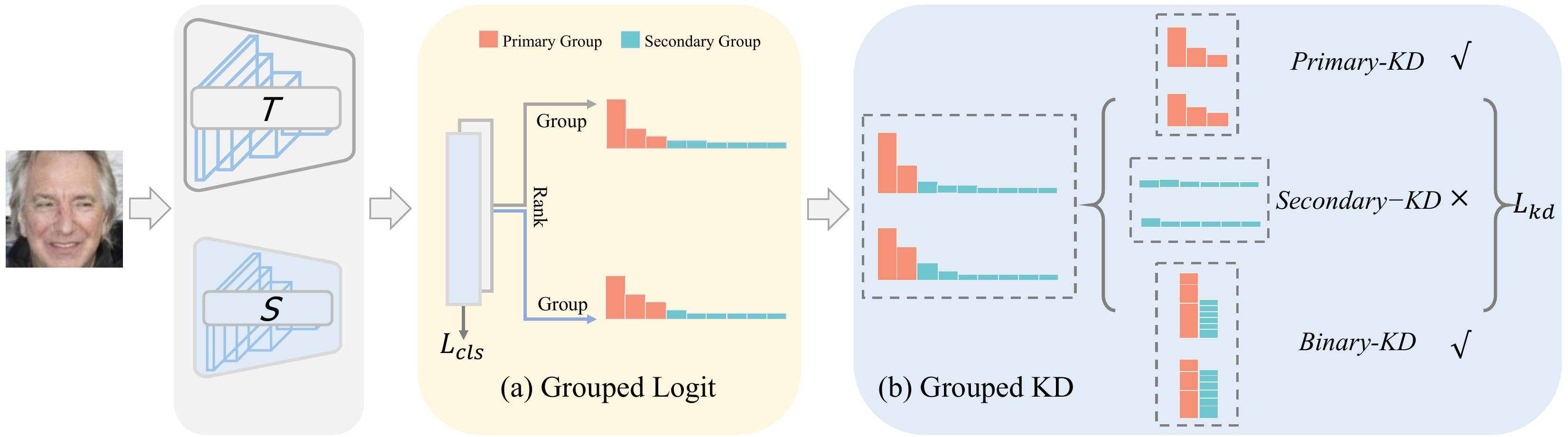}
\caption{The main framework of the proposed Grouped Knowledge Distillation (GKD). First, a facial image is fed into teacher and student networks to obtain the corresponding logits output. Then, (a) both the logits of teacher and student are ranked and partitioned into two groups, i.e., primary group and secondary group, via a cumulative probability threshold $\tau$ of student prediction. (b) With the grouped logits, the classical Knowledge Distillation loss is reorganized and divided into three parts, i.e., Primary-KD, Secondary-KD, and Binary-KD. The extensive experiments prove that (1) Primary-KD and Binary-KD are indispensable for KD, and (2) Secondary-KD is the culprit restricting KD at the bottleneck. Therefore, the proposed Grouped Knowledge Distillation (GKD) retains the Primary-KD and Binary-KD but omits Secondary-KD in the ultimate KD loss calculation. Additionally, CosFace loss is utilized as the classification loss to maintain the intra-class and inter-class relations.}
\label{fig2}
\end{figure*}
\section{Method}
In this section, we first introduce the primary group and secondary group. Then, we reformulate the classical KD Loss into three parts, i.e., Primary Knowledge Distillation, Secondary Knowledge Distillation, and Binary Knowledge Distillation, and describe the proposed Grouped Knowledged Distillation (GKD) loss. 

\subsection{Grouped Logits}
From Figure \ref{fig1}, we find that the prediction of the student network includes a long tail group with several near-zero values, which contains minor knowledge but increases the difficulty of distillation. Therefore, we seek to probe the target logits to extract the primary knowledge related to face identity, to make the distillation more viable for the student network. Specifically, we propose partitioning the original logits into two groups, i.e., primary group and secondary group. We denote the training dataset including $n$ facial images of $y$ identities as $D=\{x_t,y_t\}$, where $x_t$ refers to $t$-th sample and $y_t$ indicates its identity label. For a sample of $i$-th identity, the softened probabilities is formulated as $\mathbf{p} = \{ p_i \}, i = 1,2,\cdots,C$, where $C$ denotes the number of identities. For each $p_i \in \mathbf{p}$:
\begin{equation}
\label{eq1}
p_i = \frac{exp(z_i)}{\sum_{j=1}^C exp(z_j)},
\end{equation}
where $z_i$ indicates the logit output from the network. According to the above formulation and the results shown in Figure \ref{fig1}, there is a tail group with near-zero values in the prediction, which includes minor knowledge but increases the difficulty of distillation for the student with low model capacity.
To further explore the effects of the tail group, we propose to separate the tail group from the original logits via the cumulative probability threshold of the softened prediction, as shown in Figure \ref{fig2}. Specifically, given a ranked logits $\mathbf{z}=[z_1,z_2,\cdots,z_C]$ for sample $x$, we first obtain its ranked prediction $\mathbf{\widetilde{p}}$. Then, we partition the original logits into two groups, i.e., primary group and secondary group, and respectively denote them as $\mathbf{z}_{\Phi}$ and $\mathbf{z}_{\Psi}$, which can be formulated as follow:
\begin{equation}
\begin{aligned}
\mathbf{z}_{\Phi} &= TopK(\mathbf{z},k),\\
\mathbf{z}_{\Psi} &= \mathbf{z}\setminus \mathbf{z}_{\Phi}.
\end{aligned}
\label{eq2}
\end{equation}
We analyze that the primary group contains most of the discriminative knowledge for distillation where elements are predicted with large values from the model and vice versa for secondary group. To adaptively determine the value of $k$, we utilize a cumulative probability threshold of the student's prediction, which can be formulated as follows:
\begin{equation}
k=\mathop{\arg\min}_{k}|\sum_{i=1}^k\widetilde{p}_i-\tau|,
\label{eq3}
\end{equation}
The $\tau$ denotes the cumulative probability threshold. Generally, as the value of $\tau$ increases, the model distills more knowledge, i.e., fitting a larger proportion of target logits from the teacher network. The effects of $\tau$ are further studied in the "Effects of Threshold $\tau$" section.

\subsection{Grouped Knowledge Distillation}
The secondary group (tail group) contains minor knowledge for distillation, which we argue wastes the learning capability of the student model. To figure out the effects of secondary group, we review the formulation of the classical KD loss that refers to the Kullback-Leibler Divergence of the softened prediction between teacher and student networks. The classical KD loss can be formulated as follows:
\begin{equation}
KL(\mathbf{p}^T||\mathbf{p}^S) = \sum_{i=1}^Cp_i^T log(\frac{p_i^T}{p_i^S}),
\label{eq4}
\end{equation}
where $\mathbf{p}^T$ and $\mathbf{p}^S$ denotes the prediction of teacher and student, where the elements $p_i^T$ and $p_i^S$ are formulated as Equation \ref{eq1}. Let $\mathbf{\Phi}$, $\mathbf{\Psi}$, and $\mathbf{U}$ denote the corresponding identity sets of primary group, secondary group, and whole logits, respectively. Then, we can obtain primary group prediction $\mathbf{\hat{p}}$ and secondary group prediction $\mathbf{\check{p}}$, which are denoted as $\mathbf{\hat{p}} = \{ \hat{p}_i \}, i=1,2,\cdots, C_{\Phi}$, and $\mathbf{\check{p}} = \{ \check{p}_j \}, j = 1,2, \cdots, C_{\Psi}$, respectively. $C_{\Phi}$ and $C_{\Psi}$ represent the length of $\mathbf{\hat{p}}$ and $\mathbf{\check{p}}$, and the elements can be denoted as follows: 
\begin{equation}
\label{eq5}
\begin{aligned}
\hat{p}_i &= \frac{exp(z_i)}{\sum_{l\in \Phi}exp(z_l)}\,\,|\,i\in \Phi,\\
\check{p}_j &= \frac{exp(z_j)}{\sum_{l\in \Psi}exp(z_l)}\,\,|\,j\in \Psi.
\end{aligned}
\end{equation}
We define the notion $\mathbf{p}_b $ to represent the binary logits that indicates the knowledge distribution, which is denoted as $\mathbf{p}_{b} = [p_{\Phi},p_{\Psi}]$. Additionally, two notions $\mathbf{p}_{\Phi}$and $\mathbf{p}_{\Psi}$ are denoted to represent the binary probability of primary group and secondary group, which can be computed as follows:
\begin{equation}
\label{eq6}
\begin{aligned}
p_{\Phi} = \frac{\sum_{l\in\Phi}exp(z_l)}{\sum_{l\in\mathbf{U}}exp(z_l)},\,\,\,\,
p_{\Psi} = \frac{\sum_{l\in\Psi}exp(z_l)}{\sum_{l\in\mathbf{U}}exp(z_l)}.
\end{aligned}
\end{equation}
According to Equation \ref{eq5} and Equation \ref{eq6}, we can reorganize the formulation of KD loss in Equation \ref{eq4} with the grouped logits, which can be formulated as follows:
\begin{equation}
\label{eq7}
\begin{aligned}
&KL(\mathbf{p}^T||\mathbf{p}^S) = \sum_{i=1}^Cp_i^T log(\frac{p_i^T}{p_i^S})\\
&=\sum_{i\in\Phi}p_i^Tlog(\frac{p_i^T}{p_i^S})+\sum_{j\in\Psi}p_j^T log(\frac{p_j^T}{p_j^S})\\
&=\sum_{i\in\Phi}\hat{p_i}^Tp_{\Phi}^T log(\frac{\hat{p_i}^Tp_{\Phi}^T}{\hat{p_i}^Sp_{\Phi}^S})+\sum_{j\in\Psi}\check{p_j}^Tp_{\Psi}^T log(\frac{\check{p_j}^Tp_{\Psi}^T}{\check{p_j}^Sp_{\Psi}^S})\\
&=p_{\Phi}^T\sum_{i\in\Phi}\hat{p_i}^T log(\frac{\hat{p_i}^T}{\hat{p_i}^S})+p_{\Phi}^T log(\frac{p_{\Phi}^T}{p_{\Phi}^S})\sum_{i\in\Phi}\hat{p_i}^T\\
&+p_{\Psi}^T\sum_{j\in\Psi}\check{p_j}^T log(\frac{\check{p_j}^T}{\check{p_j}^S})+p_{\Psi}^T log(\frac{p_{\Psi}^T}{p_{\Psi}^S})\sum_{j\in\Psi}\check{p_j}^T,\\
\end{aligned}
\end{equation}
where $\mathbf{p}_{b} = [p_{\Phi},p_{\Psi}]$ represents the binary logits that indicates the knowledge distribution. From Equation \ref{eq5}, we have:
\begin{equation}
\label{eq8}
\sum_{i\in\Phi}\hat{p_i}^T = \sum_{j\in\Psi}\check{p_j}^T = 1.
\end{equation}
Therefore, the KD loss can be further formulated as follows:
\begin{equation}
\label{eq9}
\begin{aligned}
&KL(\mathbf{p}^T||\mathbf{p}^S) =p_{\Phi}^T\sum_{i\in\Phi}\hat{p_i}^T log(\frac{\hat{p_i}^T}{\hat{p_i}^S})\\
&+ p_{\Psi}^T\sum_{j\in\Psi}\check{p_j}^T log(\frac{\check{p_j}^T}{\check{p_j}^S})+[p_{\Phi}^T log(\frac{p_{\Phi}^T}{p_{\Phi}^S})+p_{\Psi}^T log(\frac{p_{\Psi}^T}{p_{\Psi}^S})]\\
&=p_{\Phi}^T\cdot KL(\mathbf{\hat{p}}^T||\mathbf{\hat{p}}^S)+p_{\Psi}^T\cdot KL(\mathbf{\check{p}}^T||\mathbf{\check{p}}^S)+KL(\mathbf{p}_b^T||\mathbf{p}_b^S).\\
\end{aligned}
\end{equation}
From the Equation \ref{eq9}, we reformulate the KD loss into three parts, i.e., Primary-KD, Secondary-KD, and Binary-KD, which can be denoted as follows:
\begin{equation}
\label{eq10}
\begin{aligned}
\mbox{Primary-KD} &= KL(\mathbf{\hat{p}}^T||\mathbf{\hat{p}}^S),\\
\mbox{Secondary-KD} &= KL(\mathbf{\check{p}}^T||\mathbf{\check{p}}^S),\\
\mbox{Binary-KD} &= KL(\mathbf{p}_b^T||\mathbf{p}_b^S).
\end{aligned}
\end{equation}
We analyze that Primary-KD refers to distilling the primary knowledge from the teacher, Secondary-KD aims at distilling minor knowledge but increases the difficulty of distillation simultaneously, and Binary-KD ensures the consistency of knowledge distribution between teacher and student. To reduce the difficulty of fitting large-scale target logits for a light-weight student network and make the distillation task more achievable, we retain the Primary-KD and Binary-KD as distillation loss, and assign proper weight to Primary-KD, which is calculated as follows:
\begin{equation}
\mathcal{L}_{kd} = \lambda_1*KL(\mathbf{\hat{p}}^T||\mathbf{\hat{p}}^S)+\lambda_2*KL(\mathbf{p}_b^T||\mathbf{p}_b^S).
\end{equation}
Additionally, we utilize ArcFace \cite{arcface} as our classification loss function to maintain intraclass differentiation and interclass aggregation, which is formulated as:
\begin{equation}
\mathcal{L}_{cls} = \frac{1}{N_i}\sum_i-log\frac{e^{s(cos(\theta_{y_i}+m,i))}}{e^{s(cos(\theta_{y_i}+m,i))}+\sum_{j \ne y_i}e^{scos(\theta_j,i)}},
\end{equation}
The overall loss function is denoted as:
\begin{equation}
\mathcal{L} = \mathcal{L}_{cls}+\mathcal{L}_{kd}.
\end{equation}
\section{Experiments}
\subsection{Dataset}
\paragraph{Training Set.} We utilize the refined MS1MV2 \cite{arcface} as our training set for fair comparisons with other SOTA methods. MS1MV2 consists of 5.8M facial images of 85K individuals.
\paragraph{Testing Set.} We evaluate our method on several popular face benchmarks, including LFW \cite{lfw}, CFP-FP \cite{cfp-fp}, CPLFW \cite{cplfw}, AgeDB \cite{agedb}, CALFW \cite{calfw}, IJB-B \cite{ijb-b}, IJB-C \cite{ijb-c}. LFW is the most commonly utilized face verification dataset, which consists of 13,233 facial images of 5,749 individuals. Cross-Age LFW (CALFW) and Cross-Pose LFW (CPLFW) databases are constructed based on the LFW database, to emphasize similar-looking challenges, cross-age and cross-pose challenges. CFP-FP database is built for facilitating large pose variation and the AgeDB-30 database is a manually collected cross-age database. The IJB-B and IJB-C are two challenging public template-based benchmarks for face recognition. The IJB-B dataset contains 1,845 subjects with 21.8K still images and 55K frames from 7,011 videos. The IJB-C dataset is a further extension of IJB-B, which contains about 3,500 identities with a total of 31,334 images and 11,7542 unconstrained video frames. MegaFace Challenge \cite{megaface} consists of the gallery set including 1M images of 690K subjects and the probe set including 100K photos of 530 individuals from FaceScrub.

\subsection{Implementation Details}
\paragraph{Data Processing.} We adopt the preprocess settings in ArcFace \cite{arcface}. The input facial images are cropped to 112$\times$112 and normalized by subtracting 127.5 and dividing by 128. For the data augmentation, we apply the horizontal flip with a probability of 50\%.
\paragraph{Training.} We utilize IResnet-50 as the teacher model that is trained by ArcFace \cite{arcface}, which is pre-trained and frozen for all face recognition model training. Additionally, we adopt two student network architectures, e.g., IResnet-18 \cite{arcface} and MobileFacenet\cite{mobilefacenet}, following the network architecture settings of EKD \cite{ekd}. To show the generality of our method, we use two neural student networks, i.e., MobileFaceNet \cite{mobilefacenet} and IResnet-18 \cite{arcface}. We set the batch size to 128 for each GPU in all experiments, and train models on 8 NVIDIA Tesla V100 (32GB) GPUs. We apply the SGD optimization method and divide the initial learning rate (0.1) at 10, 18, and 24 epochs. The momentum is set to 0.9, and the weight decay is 5e-4. The hyper-parameters $\lambda_1$ and $\lambda_2$ are set to 8.0 and 1.0, respectively. For ArcFace, we follow the common setting with s = 64 and margin m = 0.5. 
\paragraph{Testing.}
We follow the evaluation protocol and network settings proposed by EKD \cite{ekd} to report the performance on LFW, CFP-FP, CPLFW, AgeDB, CALFW, IJB-B, IJB-C, and MegaFace Challenge.

\subsection{Ablation Study}
In this section, we first conduct the ablation experiments on Primary-KD, Secondary-KD, and Binary-KD. Then, we explore the effects of different cumulative probability threshold $\tau$ and that of hyper-parameters $\lambda_1$ and $\lambda_2$. Additionally, we investigate the generalization capability of our method with different student network structures, i.e., IResNet-18 and MobileFacenet. All the experiments are evaluated on five popular face benchmarks, i.e., LFW, CFP-FP, CPLFW, AgeDB, and CALFW. 
\begin{table}[htb]
\small
    \centering
\resizebox{\columnwidth}{!}{%

    \begin{tabular}{@{}cccccccc@{}}

    \toprule
        \textbf{P-KD} & \textbf{B-KD} & \textbf{S-KD} & \textbf{CFP-FP} & \textbf{CPLFW} & \textbf{AgeDB} & \textbf{CALFW}  \\ \midrule
         $\checkmark$ & $\checkmark$ & $\checkmark$  & 91.71 & 87.85 & 95.93 & 95.03  \\  
         $\checkmark$ &  &  & 94.08 & 90.56 & 96.73 & 95.45  \\ 
	    $\checkmark$ & $\checkmark$ &  & \textbf{94.35} & \textbf{90.86} & \textbf{97.25} & \textbf{95.78}  \\ \bottomrule 
    \end{tabular}%
}
\caption{Ablation experiments of Primary-KD (P-KD), Secondary-KD (S-KD), and Binary-KD (B-KD).}
\label{tab:1}
\end{table}
\paragraph{Ablation on GKD.}
We divide the classical KD loss into three parts, i.e., Primary-KD, Secondary-KD, and Binary-KD. To investigate their effects, we conduct the ablation experiments on them, and we can see from Table \ref{tab:1} that the classical KD loss achieves inferior performance, and the recognition performance achieves a significant improvement on five testing sets when we omit the Secondary-KD while retaining the Primary-KD and Secondary-KD. This phenomenon indicates the effectiveness of distilling primary knowledge and attributes Secondary-KD as the culprit for the bottleneck of knowledge distillation. We analyze that liberalizing the requirements for the student network to fit secondary knowledge can reduce the difficulty of distillation, thus bridging the performance gap between teacher and student. Additionally, Binary-KD also brings a slight performance improvement, since it specifies the knowledge distribution of model output and keeps consistent knowledge distribution between teacher and student networks to avoid the overfitting of the student model. From the results shown in Table \ref{tab:1}, we experimentally find that (1) Primary-KD and Binary-KD are indispensable for KD, and (2) Secondary-KD is the culprit restricting KD at the bottleneck. Therefore, we propose a Grouped Knowledge Distillation (GKD), which retains the Primary-KD and Binary-KD but omits Secondary-KD in the ultimate KD loss calculation.
\begin{table}[htb]
    \centering
\resizebox{\columnwidth}{!}{%
    \begin{tabular}{@{}cccccc@{}}
    \toprule
        \textbf{Method} & \textbf{LFW} & \textbf{CFP-FP} & \textbf{CPLFW} & \textbf{AgeDB} & \textbf{CALFW} \\ \midrule
        Student & 99.52 & 91.66 & 87.93 & 95.82 & 95.12 \\ 
        KD ($\mathbf{\tau}=1$) & 99.50 & 91.71 & 87.85 & 95.93 & 95.03 \\ \midrule
        $\mathbf{\tau}=0.99$ & 99.46 & 92.15 & 87.98 & 96.16 & 95.41 \\ 
        $\mathbf{\tau}=0.97$ & 99.53 & 93.81 & 90.65 & 96.90 & 95.53 \\ 
        $\mathbf{\tau}=0.95$ & 99.55 & 94.25 & 90.43 & 96.88 & 95.60 \\ 
        $\mathbf{\tau=0.93}$ & \textbf{99.61} & \textbf{94.35} & \textbf{90.86} & \textbf{97.25} & \textbf{95.78} \\ 
        $\mathbf{\tau}=0.91$ & 99.52 & 94.00 & 89.95 & 96.93 & 95.65 \\ \bottomrule
    \end{tabular}%
}
\caption{Extensive ablation on cumulative threshold $\tau$.}
\label{tab2}
\end{table}
\paragraph{Effects of Threshold $\mathbf{\tau}$.}
\label{tau}
As described in Equation \ref{eq2}, we partition the original logits into two groups, i.e., primary group and secondary group, via a cumulative probability threshold $\tau$. To explore the effects of $\tau$, we conduct comparisons for different thresholds. Generally, a larger threshold indicates more knowledge for a student to fit and refers to KD loss when $\tau=1$. Moreover, we can see from Table \ref{tab2} that the model trained with $\tau=0.99$ has a similar performance in comparison to the classical KD loss. As $\tau$ decreases, less knowledge is required to be fitted for the student network, corresponding to achievable distillation tasks and better performance. Additionally, the best recognition performance comes when $\tau$ equals $0.93$, and we set $\tau = 0.93$ as the default value of $\tau$ in our subsequent ablation and comparison experiments.
\begin{table}[htb]
    \centering
\resizebox{\columnwidth}{!}{%
    \begin{tabular}{@{}cccccc@{}}
    \toprule
        \textbf{Method} & \textbf{LFW} & \textbf{CFP-FP} & \textbf{CPLFW} & \textbf{AgeDB} & \textbf{CALFW}  \\ \midrule
        Student & 99.52 & 91.66 & 87.93 & 95.82 & 95.12  \\ \midrule
        $\mathbf{\lambda_1} =p_{\Phi}^T $ & 99.58 & 94.25 & 89.85 & 97.21 & 95.65  \\ 
        $\mathbf{\lambda_1} =2$ & 99.46 & 94.07 & 90.63 & 96.80 & 95.60 \\ 
        $\mathbf{\lambda_1} =4$ & 99.52 & 94.23 & 90.75 & 96.95 & 95.50 \\ 
        $\mathbf{\lambda_1}=8 $ & \textbf{99.61} & \textbf{94.35} & \textbf{90.86} & \textbf{97.25} & \textbf{95.78}  \\ 
        $\mathbf{\lambda_1} =10$ & 99.48 & 94.03 & 90.58 & 97.22 & 95.66  \\ \midrule\midrule
		$\mathbf{\lambda_2} =0$ & 99.51 & 94.08 & 90.56 & 96.73 & 95.45  \\ 
	    $\mathbf{\lambda_2} =1$ & 99.61 & \textbf{94.35} & \textbf{90.86} & 97.25 & \textbf{95.78}  \\ 
	    $\mathbf{\lambda_2} =2$ & \textbf{99.64} & 94.23 & 90.68 & \textbf{97.28} & 95.69  \\
	\bottomrule
    \end{tabular}%
}
\caption{Ablation on different $\lambda_1$ and $\lambda_2$, which correspond to weights for Primary-KD and Secondary-KD, respectively.}
\label{tab3}
\end{table}
\paragraph{Effects of $\lambda_1$ and $\lambda_2$.}
The table \ref{tab3} reports the recognition accuracy of student networks with different $\lambda_1$ and $\lambda_2$. First, we can find the default value of $\lambda_1=\mathbf{p}_{\Phi}^T$ can bring reasonable performance improvement. Then, we conduct comparisons among different values of it and find the model achieves better performance when $\lambda_1$ increases. We analyze that the default weight $\mathbf{p}_{\Phi}^T$ cannot highlight the significance of Primary-KD, which can be enhanced by a larger weight. Additionally, we find that different weights do not introduce significant performance improvement since we discard Secondary-KD, which disguisedly enhances Primary-KD learning. The best performance is obtained when $\lambda_1=8$. Moreover, based on the setting ($\lambda_1=8$), we perform the ablation on different values of $\lambda_2$. From Table \ref{tab3}, we find that different values of $\lambda_2$ have a minor effect on the final result, as long as Binary-KD is introduced. The best performance achieves when $\lambda_2 = 1.0$.
\begin{table}[htb]
\small
    \centering
    \begin{tabular}{@{}ccc@{}}
    \toprule
        \multirow{2}{*}{\textbf{Method}} & \multicolumn{2}{c}{\textbf{IJB-C}}  \\ \cmidrule(r){2-3}
& 1e-4 & 1e-5 \\ \midrule
	  MobileFaceNet & 89.13 & 81.65  \\
MobileFaceNet+EKD &90.48 & 84.00 \\
 MobileFaceNet+Ours & \textbf{94.34} & \textbf{91.01}  \\ \midrule
        IR18 & 91.96 & 86.01  \\ 
IR18+EKD & 92.74 & 88.84 \\
        IR18+Ours & \textbf{94.93} & \textbf{92.29}  \\ \bottomrule
    \end{tabular}%
\caption{Ablation studies of student networks, which involves IResNet-18 and MobileFacenet. TPR@FPR=1e-4 and TPR@FPR=1e-5 on IJB-C are reported.}
\label{tab4}
\end{table}
\paragraph{Effects of Student Network Architecture.}
We investigate the generalization capability of our method for different student network structures. Table \ref{tab4} shows the results of two structures, i.e., IResNet-18 and MobileFaceNet. Although the performance improvement on the two network structures is different, our method generally performs better than directly training the student network from scratch. Additionally, we keep the consistent student network (IResNet-18 and MobileFacenet) and conduct a comparison with EKD \cite{ekd}. The results of Table \ref{tab4} show that our method improves the student model.
\begin{table*}[htb]
\small
    \centering
\resizebox{\textwidth}{!}{%
    \begin{tabular}{@{}ccccccccccccc@{}}
    \toprule
        \multirow{2}{*}{\textbf{Method}} & \textbf{\textbf{IJB-C}} & \textbf{\textbf{IJB-B}}& \textbf{\textbf{MegaFace}} & \textbf{LFW} & \textbf{CFP-FP} & \textbf{CPLFW} & \textbf{AgeDB} & \textbf{CALFW} \\ \cmidrule(r){2-9}
&1e-4/1e-5 & 1e-4/1e-5 & Id/Ver & ACC & ACC & ACC & ACC & ACC \\ \midrule 
        Teacher-IR50 (upper bound)& 95.16/92.66 & 93.45/88.65 &98.14/98.34& 99.80 & 97.63 & 92.50 & 97.92 & 96.05  \\ 
        MobileFaceNet (student)& 89.13/81.65 & 87.07/74.63 &90.91/92.71& 99.52 & 91.66 & 87.93 & 95.82 & 95.12  \\ \midrule
        FitNet \cite{fitnet}& 87.76/73.71 & 86.35/70.19 &91.16/92.34& 99.47 & 91.30 & 88.30 & 96.18 & 95.12  \\ 
        KD \cite{kd} & 88.37/80.39 & 86.08/74.30 &90.40/92.00& 99.50 & 91.71 & 87.85 & 95.93 & 95.03  \\ 
        Dark \cite{darkrank} & 89.28/81.62 & 86.76/73.75 &90.76/92.41& 99.55 & 91.84 & 87.77 & 95.60 & 95.07  \\ 
        SP \cite{sp} & 88.43/78.13 & 86.34/72.85 &91.25/92.41& 99.53 & 92.33 & 88.45 & 96.17 & 95.07  \\ 
        CCKD \cite{cckd} & 87.99/78.75 & 85.63/72.38 &91.17/92.76& 99.47 & 91.90 & 88.48 & 95.83 & 95.22  \\
        RKD \cite{cckd} & 89.65/83.21 & 87.27/75.17 &91.44/92.92& 99.58 & 92.13 & 87.97 & 96.18 & 95.25  \\ 
        ShrinkTeaNet \cite{shrinkteanet} & 87.80/79.78 & 85.31/75.23 &90.73/92.32& 99.47 & 91.97 & 88.52 & 96.00 & 94.98  \\ 
        Triplet \cite{tripletdistillation}& 84.57/76.65 & 81.88/70.51 &86.52/88.75& 99.55 & 93.14 & 88.03 & 95.33 & 94.97  \\ \
        Margin \cite{margindistillation} & 85.71/75.00 & 82.97/66.25 &91.70/92.96& 99.61 & 92.01 & 88.03 & 96.55 & 95.13  \\ 
        EKD \cite{ekd}& 90.48/84.00 & 88.35/76.60 &91.02/93.08& 99.60 & 94.33 & 89.35 & 96.48 & 95.37  \\ 
        \textbf{Ours} & \textbf{94.34}/\textbf{91.01} & \textbf{92.52}/\textbf{86.06} &\textbf{95.48}\textbf{/}\textbf{95.87}& \textbf{99.61} & \textbf{94.35} & \textbf{90.86} & \textbf{97.25} & \textbf{95.78}  \\ \bottomrule
    \end{tabular}%
}
\caption{Verification comparison with SOTA methods on LFW, two pose benchmarks: CFP-FP and CPLFW, two age benchmarks: AgeDB and CALFW, and large-scale benchmarks: IJB-B, IJB-C and MegaFace.}
\label{tab6}
\end{table*}
\subsection{Comparison with SOTA}
In this section, we compare our proposed GKD with several Sate-Of-The-Art (SOTA) knowledge distillation methods, including the methods proposed for other tasks (KD \cite{kd}, FitNet \cite{fitnet}, DarkRank \cite{darkrank}, SP \cite{sp}, CCKD \cite{cckd} and RKD \cite{cckd}), and methods designed for face recognition (ShrinkTeaNet \cite{shrinkteanet}, Triplet Distillation \cite{tripletdistillation}, MarginDistillation \cite{margindistillation} and EKD \cite{ekd}). The results of six methods designed for other tasks are reproduced and reported on face recognition benchmarks. For the methods designed for face recognition, we replicate the results from EKD.
\paragraph{Results on LFW, CFP-FP, CPLFW, AgeDB, and CALFW.}
To evaluate the superiority of our proposed GKD, we first evaluate the recognition performance on five widely-used face benchmarks to compare with other SOTA methods. As shown in Table \ref{tab6}, most of the knowledge distillation methods are better than the student network that is directly trained from scratch (i.e., MobileFaceNet), but the performance improvement is limited. Additionally, feature-based distillation, e.g., FitNet and RKD, seem to show better performance than the logits-based methods (KD), in comparison with all the competitors, while performing inferior to MarginDistillation. By contrast, the SOTA method EKD brings significant improvement in comparison to other methods. Compared with the feature-based EKD, our method follow the logits distillation and achieves explicit improvements on five face benchmarks, which further bridges the performance gap by alleviating the difficulty of distillation.   
\paragraph{Results on IJB-B and IJB-C.}
In Table \ref{tab6}, we extensively conduct the comparisons of the 1:1 verification (TPR@FPR=1e-4 and TPR@FPR=1e-5) with the previous SOTA methods on the IJB-B and IJB-C. Surprisingly, the results of our proposed method achieve a significant verification performance improvement, which is different from the results in the small test datasets. Additionally, most of the knowledge distillation methods bring little performance improvement or are even worse than the baseline on these two large-scale test datasets. Compared with previous methods, both RKD and EKD bring supervising improvements, but our method further improves $3.86\%$ and $7.01\%$ verification performance on IJB-C and brings $4.17\%$ and $9.46\%$ improvements on IJB-B (TPR@FPR=1e-4 and TPR@FPR=1e-5), which indicates the effectiveness of our method in the large-scale verification scenarios. 
\paragraph{Results on MegaFace.}
We test our model on MegaFace Challenge 1 using FaceScrub as the probe set. “Id” refers to Rank-1, and Ver refers to TPR@FPR=1e-6. As shown in Table \ref{tab6}, the proposed GKD outperforms other methods on MegaFace.
\begin{table}[htb]
    \centering
\resizebox{\columnwidth}{!}{%
    \begin{tabular}{@{}cccccc@{}}
    \toprule
        \textbf{Method} & Student & FitNet & KD & EKD & GKD  \\ \midrule
        \textbf{Training Time(s)}& 0.088 & 0.120 & 0.116& 0.133 & 0.121 \\
	\bottomrule
    \end{tabular}%
}
\caption{Training time comparison for each batch under the same experimental setting.}
\label{tab-time}
\end{table}
\paragraph{Time Complexity.}
As shown in Table \ref{tab-time}, we assess the training time of students including 1) student (w/o distillation loss), 2) KD, 3) FitNet, 4) EKD, and 5) our method for each batch under the same experimental setting. Specifically, we conduct 4000 complete iterations including forward and backward propagation on one NVIDIA Tesla-V100, and calculate the average as the training time. The batch size is set to 8 and the Pytorch version is 1.7.1. 
\begin{table}[htb]
    \centering
\resizebox{\columnwidth}{!}{%
    \begin{tabular}{@{}ccccccc@{}}
    \toprule
        \multirow{2}{*}{\textbf{Method}} &\textbf{\textbf{CALFW}} & \textbf{\textbf{CPLFW}}  & \textbf{\textbf{AgeDB}} \\ \cmidrule(r){2-4}
        ~ & M-M/M-N & M-M/M-N & M-M/M-N \\ \midrule
        Teacher (IR100) & 95.80/95.58 & 91.68/92.51 & 96.30/97.00  \\ 
        MobileNetV3-L & 92.65/93.33 & 85.91/87.23 & 90.30/93.46 \\ 
        MobileNetV3-L+Ours & \textbf{94.41}/\textbf{94.66} & \textbf{89.46}/\textbf{90.48} & \textbf{94.50}/\textbf{95.83}  \\ \bottomrule
    \end{tabular}%
}
\caption{Ablation studies on Masked Face Recognition challenge with different student-teacher network settings, which involves in MobileNetV3-large300 (Student) and IResNet-100 (Teacher). The testing scenarios M-M and M-N indicate mask vs. mask and mask vs. non-mask, respectively.}
\label{tab5}
\end{table}
\subsection{Generalization on Masked Face Recognition}
In this section, we study the generalization of our method to other recognition tasks, e.g., Masked Face Recognition. Knowledge distillation is a general method to bridge the performance gap between teacher and student networks, and we think its effectiveness should be verified on other tasks, e.g., Masked Face Recognition (MFR). We conduct ablation experiments on MFR with different network settings, i.e., MobileNetV3-large (Student) \cite{mobilenetv3} and IResNet-100 (Teacher) \cite{arcface}. For the training dataset, we adopt the FaceX-Zoo \cite{facexzoo} to generate the masked data from MS1M \cite{arcface}, which consists of approximately 10M images of 9.3K identities. We utilize CosFace \cite{cosface} loss with margin $m=0.4$ and scale $s=64$ as the loss function. There are two scenarios for MFR, i.e., Mask vs. Mask (M-M), and Mask vs. Non-mask (M-N). For the testing sets and evaluation protocol, we keep consistent with MaskInv \cite{maskinv}. From Table \ref{tab5}, we can see that our method outperforms the baseline method, which demonstrates the generalization of our model on masked face recognition.
\section{Conclusion}
This paper proposes Grouped Knowledge Distillation (GKD) to probe the target logits to extract the primary knowledge that is related to face identity, and discard the others, to make the distillation more achievable for the student network. Specifically, there is a tail group that has near-zero values in the prediction, including minor knowledge of distillation. Therefore, we first partition the logits into two groups, i.e., primary group and secondary group, via the cumulative probability of the softened prediction. Then, we reorganize the distillation loss into three parts, i.e., Primary-KD, Secondary-KD, and Binary-KD. We experimentally found that Primary-KD and Binary-KD are indispensable for KD, and Secondary KD is the culprit restricting KD at the bottleneck. Extensive experimental results on popular face recognition benchmarks demonstrate the superiority of proposed GKD over state-of-the-art methods. Moreover, the experiments conducted on masked face recognition tasks demonstrate the generalization of our method as well.

\section{Acknowledgments}
This work was supported in part by the National Key Research \& Development Program (No. 2020YFC2003901), Chinese National Natural Science Foundation Projects (61876178, 62276254, 62176256, 62106264, 62206280, 61871378, U2003111), the Youth Innovation Promotion Association CAS (\#Y2021131), Defense Industrial Technology Development Program (Grant JCKY2021906A001) and the InnoHK program. 

\bibliography{aaai23}

\end{document}